\documentclass[11pt,a4paper]{article}
\usepackage[hyperref]{acl2019}
\usepackage{times}
\usepackage{latexsym}
\usepackage{graphicx}
\usepackage{url}
\usepackage{mathtools}
\usepackage{amssymb}
\usepackage{amsthm}
\usepackage{multirow}
\usepackage{subfigure}
\usepackage{booktabs}
\usepackage{algorithm}
\usepackage{algorithmic}
\usepackage{bm}
\newcommand{\red}[1]{\textcolor{red}{#1}}
\newcommand{\blue}[1]{\textcolor{blue}{#1}}

\usepackage{eqparbox}

\newcommand{\tsc}[1]{\textsuperscript{#1}}
\usepackage{url}

\aclfinalcopy 
\def\aclpaperid{1128} 

\title{A Hierarchical Reinforced Sequence Operation Method for \\ Unsupervised Text Style Transfer}

\author{Chen Wu\tsc{1}\thanks{\ \ Equal Contributions.}, Xuancheng Ren\tsc{2}\footnotemark[1], Fuli Luo\tsc{2}, Xu Sun\tsc{2,3}\\
  \tsc{1}Department of Foreign Languages and Literatures, Tsinghua University\\
  \tsc{2}MOE Key Laboratory of Computational Linguistics, School of EECS, Peking University\\
  \tsc{3}Center for Data Science, Beijing Institute of Big Data Research, Peking University\\
   {\tt wu-c16@mails.tsinghua.edu.cn}\\
   {\tt \{renxc, luofuli, xusun\}@pku.edu.cn} \\}
   
\date{}

\begin{document}
\maketitle
\begin{abstract}
  Unsupervised text style transfer aims to alter text styles while preserving the content, without aligned data for supervision. Existing seq2seq methods face three challenges: 1) the transfer is weakly interpretable, 2) generated outputs struggle in content preservation, and 3) the trade-off between content and style is intractable. To address these challenges, we propose a hierarchical reinforced sequence operation method, named \textbf{Point-Then-Operate} (\textbf{PTO}), which consists of a high-level agent that proposes operation positions and a low-level agent that alters the sentence. We provide comprehensive training objectives to control the fluency, style, and content of the outputs and a mask-based inference algorithm that allows for multi-step revision based on the single-step trained agents. Experimental results on two text style transfer datasets show that our method significantly outperforms recent methods and effectively addresses the aforementioned challenges. \footnote{\ Our code is available at \url{https://github.com/ChenWu98/Point-Then-Operate}.}
\end{abstract}

\section{Introduction}
\label{introduction}
Text style transfer aims to convert a sentence of one style into another while preserving the style-independent content \cite{shen2017style, fu2018style}. In most cases, aligned sentences are not available, which requires learning from non-aligned data. Previous work mainly learns disentangled content and style representations using seq2seq \cite{sutskever2014sequence} models and decomposes the transfer into \textit{neutralization} and \textit{stylization} steps. Although impressive results have been achieved, three challenges remain: 1) the interpretability of the transfer procedure is still weak in seq2seq models, 2) generated sentences are usually highly stylized with poor content preservation, and 3) the trade-off between content preservation and style polarity is intractable. 
\begin{figure}[t]
\centering
\includegraphics[width=1.0\linewidth]{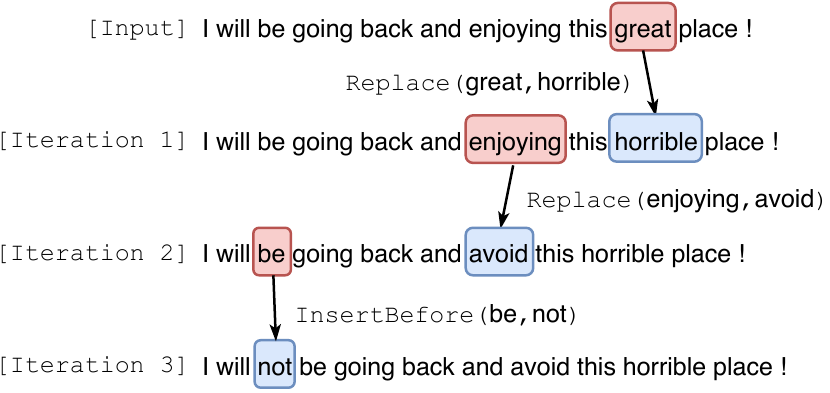}
\caption{Our proposed Point-Then-Operate (PTO) applied to a real test sample. A high-level agent (red squares) iteratively proposes operation positions, and a low-level agent (arrows) alters the sentence based on the high-level proposals. Compared with seq2seq methods, PTO is more interpretable and better preserves style-independent contents.}
\label{fig:example}
\end{figure}

To address these challenges, we propose a sequence operation-based method within the hierarchical reinforcement learning (HRL) framework, named \textbf{Point-Then-Operate} (\textbf{PTO}). It consists of a hierarchy of a high-level agent that proposes operation positions and a low-level agent that alters the sentence based on high-level proposals. We propose a policy-based training algorithm to model the key aspects in text style transfer, i.e., fluency, style polarity, and content preservation. For fluency, we use a \textit{language model reward}; for style polarity, we introduce a \textit{classification confidence reward} and an auxiliary classification task; for content preservation, we adopt a \textit{reconstruction reward} and a self-supervised \textit{reconstruction loss}. We introduce a mask-based inference algorithm that applies multi-step sequence operations to the input sentence, allowing for single-step training which is more stable. Figure~\ref{fig:example} shows an example of our method applied to a real test sample from Yelp. 

Compared with existing seq2seq methods, our sequence operation method has three merits. 1) \textit{Interpretability}: our method explicitly models \textit{where} and \textit{how} to transfer. 2) \textit{Content preservation}: sequence operations are targeted at stylized parts; thus, style-independent content can be better preserved. 3) \textit{Controllable trade-off}: the trade-off between content preservation and style polarity could be tuned in our method. Specifically, we tune it by biasing the number of operation steps.

We conduct extensive experiments on two text style transfer datasets, i.e., Yelp and Amazon. We show that our proposed method outperforms recent methods and that it addresses the challenges of existing seq2seq methods. The contributions of this paper are: 
\begin{itemize}
    \item We propose a sequence operation method, i.e., Point-Then-Operate, for unsupervised text style transfer. The transfer procedure is modeled as explicit revisions on the input sentences, which improves interpretability, content preservation, and controllable style-content trade-off.
    \item The method is interpreted and trained in the HRL framework with a high-level agent that proposes operation positions and a low-level agent that applies explicit operations. We design comprehensive learning objectives to capture three important aspects of text style transfer and propose a mask-based inference algorithm that allows for multi-step revision based on the single-step trained agents.
    \item Experiments on Yelp and Amazon show that our method significantly improves BLEU, fluency, and content preservation compared with recent methods and effectively addresses the aforementioned challenges.
\end{itemize}

\section{Related Work}
\paragraph{Text Style Transfer} Most work on text style transfer learns disentangled representations of style and content. We categorize them based on how they represent content. \textit{Hidden vector approaches} represent content as hidden vectors, e.g., \citet{hu2017toward} adversarially incorporate a VAE and a style classifier; \citet{shen2017style} propose a cross-aligned AE that adversarially aligns the hidden states of the decoder; \citet{fu2018style} design a multi-decoder model and a style-embedding model for better style representations; \citet{yang2018unsupervised} use language models as style discriminators; \citet{john2018disentangled} utilize bag-of-words prediction for better disentanglement of style and content. \textit{Deletion approaches} represent content as the input sentence with stylized words deleted, e.g., \citet{li2018delete} delete stylized n-grams based on corpus-level statistics and stylize it based on similar, retrieved sentences; \citet{xu2018unpaired} jointly train a neutralization module and a stylization module the with reinforcement learning; \citet{zhang2018learning} facilitate the stylization step with a learned sentiment memory. 

As far as we know, there are two work that avoid disentangled representations. \citet{zhang2018style} construct a pseudo-aligned dataset with an SMT model and then learn two NMT models jointly and iteratively. A concurrent work, \citet{Luo19Dual}, propose to learn two dual seq2seq models between two styles via reinforcement learning, without disentangling style and content.

\paragraph{Sequence Operation Methods} Our work is also closely related to sequence operation methods, which are widely used in SMT \cite{DBLP:conf/acl/DurraniSF11, DBLP:journals/coling/DurraniSFKS15, DBLP:conf/wmt/PalZG16} and starts to attract attention in NMT \cite{DBLP:conf/emnlp/StahlbergSB18}. Compared with methods based on seq2seq models, sequence operation methods are inherently more interpretable \cite{DBLP:conf/emnlp/StahlbergSB18}. Notably, our method is revision-based, i.e., it operates directly on the input sentence and does not generate from scratch as in machine translation systems.

\paragraph{Hierarchical Reinforcement Learning} In this work, we adopt the Options Framework \cite{SuttonPS99} in HRL, in which a high-level agent learns to determine more abstract \textit{options} and a low-level agent learns to take less abstract \textit{actions} given the option. Recent work has shown that HRL is effective in various tasks, e.g., Atari games \cite{KulkarniNST16}, relation classification \cite{FengHZYZ18}, relation extraction \cite{takanobu2019hierarchical}, and video captioning \cite{WangCWWW18}.

\section{Formulation}
We start by formalizing the problem of our interest. Given two non-aligned sets of sentences $\mathcal{X}_{1}=\{\bm{x}^{(1) }_{1}, \cdots, \bm{x}^{(n) }_{1}\}$ of style $s_{1}$ and $\mathcal{X}_{2}=\{\bm{x}^{(1)}_{2}, \cdots, \bm{x}^{(m)}_{2}\}$ of style $s_{2}$. Unsupervised text style transfer aims to learn two conditional distributions $p(\bm{x}_{1 \rightarrow 2}|\bm{x}_{1})$ and $p(\bm{x}_{2 \rightarrow 1}|\bm{x}_{2})$ which alter the style of a sentence and preserve the style-independent content. However, defining \textit{content} is not trivial. Different from previous text style transfer methods that \textit{explicitly} model contents with disentangled representations, we \textit{implicitly} model content with reconstruction, similar to the idea proposed adopted in CycleGAN \cite{zhu2017unpaired}. Given the discreteness nature of natural language texts, we use sequence operations to approximate $p(\bm{x}_{1 \rightarrow 2}|\bm{x}_{1})$ and $p(\bm{x}_{2 \rightarrow 1}|\bm{x}_{2})$. In our notations, $\bm{x}_{1 \rightarrow 2}$ and $\bm{x}_{2 \rightarrow 1}$ are \textit{transferred sentences}, which are the outputs of a text style transfer system; $\hat{\bm{x}}_{2}$ and $\hat{\bm{x}}_{1}$ are \textit{operated sentences}, which are not necessarily fully transferred.

\section{Our Approach}
Our proposed sequence operation-based method, \textbf{Point-Then-Operate} (\textbf{PTO}), decomposes style transfer into two steps: 1) finding \textit{where} to transfer and 2) determining \textit{how} to transfer. It could be naturally formulated as an HRL problem, in which a high-level agent (i.e., \textit{pointer}) proposes operation positions and a low-level agent (i.e., \textit{operators}) alters the sentence based on high-level proposals. 

In this section, we first briefly review the Options Framework in HRL. Then we introduce the proposed pointer module (\S\ref{pointer}) and operator modules (\S\ref{operators}). The training algorithm is in \S\ref{one-step}, in which two extrinsic rewards, an intrinsic reward, and a self-supervised loss are proposed for fluency, style polarity, and content preservation. The inference algorithm is in \S\ref{multi-step}, in which a mask mechanism is proposed to iteratively and dynamically apply sequence operations to the input.
\subsection{Review: The Options Framework in HRL}
The Options framework \cite{SuttonPS99} is a well-known formulation in HRL. We denote the state space as $\mathcal{S}$; the option space, $\mathcal{O}$; the action space, $\mathcal{A}$. The high-level agent learns a stochastic policy $\mu: \mathcal{S} \times \mathcal{O} \rightarrow [0, 1]$. The low-level agent learns a stochastic policy $\pi_{o}: \mathcal{S} \times \mathcal{A} \rightarrow [0, 1]$, conditioned on an option $o \in \mathcal{O}$. Additionally, each option $o \in \mathcal{O}$ has a low-level stochastic termination condition $\beta_{o}: S \rightarrow [0, 1]$ which indicates whether the current option should end. In each episode, the high-level agent executes a trajectory $(o_{1}, \cdots, o_{L})$ based on $\mu$; once an option $o_{t}$ is sampled, the low-level agent executes a trajectory $(a^{1}_{t}, \cdots, a^{l_{t}}_{t})$ based on $\pi_{o_{t}}$, where $l_{t}$ is dependent on $\beta_{o_{t}}$. Intuitively, the flattened trajectory for one episode is $(o_{1}, a^{1}_{1}, \cdots, a^{l_{1}}_{1}, \cdots, o_{L}, a^{1}_{L}, \cdots, a^{l_{L}}_{L})$. 

\subsection{High-Level Agent: Pointer}
\label{pointer}
The high-level policy $\mu$ aims to propose operation positions; thus, we model it as an attention-based \cite{bahdanau2014neural} \textit{pointer} network, which assigns normalized probability to each position.
\paragraph{Option} Given a sentence $\bm{x} = \{x_{1}, \cdots, x_{T}\}$, the option space is $\mathcal{O} = \{1, \cdots, T\}$. Note that $T$ changes within an episode, since operations may change the length of a sentence.
\paragraph{State} The state is represented by the sentence representation $\bm{h}_{T}$ and each position representation $\bm{h}_{i}$, where $\{\bm{h}_{1}, \cdots, \bm{h}_{T}\}$ is mapped from the sentence $\bm{x}$ by a bi-LSTM encoder.
\paragraph{Policy} We adopt an attention-based policy $\mu$:
\begin{equation}
\label{eq:ptr}
    \mu(i|\bm{x})=\frac{\exp(a(\bm{h}_{T}, \bm{h}_{i}))}{\sum_{t=1}^{T}\exp(a(\bm{h}_{T}, \bm{h}_{t}))}
\end{equation}
where $a(\cdot, \cdot)$ is the scoring function for attention, and $i \in \{1, \cdots, T\}$ denotes each position in the intput sentence.

\subsection{Low-Level Agent: Operators} 
\label{operators}
The low-level policy $\pi$ alters the sentence around the position $i$ (i.e., option) sampled from $\mu$. We restrict the operations to those listed in Table~\ref{tab:operators}. Note that these operations are complete to generate all natural language sentences in multiple steps. 
\begin{table}[t]
\centering
\small
\begin{tabular}{@{}ll@{}}
\toprule
\bf Module & \bf Operation \\
\midrule
$\textrm{IF}_{\phi_{1}}$ & \textbf{I}nsert a word $\hat{w}$ in \textbf{F}ront of the position \\
$\textrm{IB}_{\phi_{2}}$ & \textbf{I}nsert a word $\hat{w}$ \textbf{B}ehind the position \\
$\textrm{Rep}_{\phi_{3}}$ & \textbf{Rep}lace it with another word $\hat{w}$ \\
$\textrm{DC}$ & \textbf{D}elete the \textbf{C}urrent word \\
$\textrm{DF}$ & \textbf{D}elete the word in \textbf{F}ront of the position \\
$\textrm{DB}$ & \textbf{D}elete the word \textbf{B}ehind the position \\
$\textrm{Skip}$ & Do not change anything \\
\bottomrule
\end{tabular}
\caption{\label{tab:operators} Operator modules. Parameters $\phi_{1}$, $\phi_{2}$, and $\phi_{3}$ are meant to generate their corresponding $\hat{w}$.}
\end{table}
\paragraph{Action} Given the sentence $\bm{x} = \{x_{1}, \cdots, x_{T}\}$ and the operation position $i$, the action of the low-level agent can be decomposed into two step, i.e.,
\begin{enumerate}
    \item \textit{Operator selection}. Select an operator module from Table~\ref{tab:operators}.
    \item \textit{Word generation }(optional). Generates a word, if necessary as is specified in Table~\ref{tab:operators}.
\end{enumerate}
\paragraph{State} Compared with the high-level agent, our low-level agent focuses on features that are more local. We map $\bm{x}$ to $\{\bm{h}_{1}, \cdots, \bm{h}_{T}\}$\footnote{We reuse $\bm{h}$ and $\bm{W}$ notations for all modules for brevity.} through a bi-LSTM encoder and take $\bm{h}_{i}$ as the state representation.
\paragraph{Low-Level Termination Condition} Different from the original Options Framework in which a stochastic termination condition $\beta_{o}$ is learned, we adopt a deterministic termination condition: the low-level agent takes \textit{one} action in each option and terminates, which makes training easier and more stable. Notably, it does not harm the expressiveness of our method, since multiple options can be executed.
\paragraph{Policy for Operator Selection} For training, we adopt a \textit{uniform} policy for operator selection, i.e., we uniformly sample an operator module from Table~\ref{tab:operators}. In preliminary experiments, we explored a learned policy for operator selection. However, we observed that the learned policy quickly collapses to a nearly deterministic choice of $\textrm{Rep}_{\phi_{3}}$. Our explanation is that, in many cases, replacing a stylized word is the optimal choice for style transfer. Thus, the uniform policy assures that all operators are trained on sufficient and diversified data. For inference, we adopt a heuristic policy based on fluency and style polarity, detailed in \S\ref{choice-of-operator-modules}.
\paragraph{Policy for Word Generation} As shown in Table~\ref{tab:operators}, three operators are parameterized, which are burdened with the task of generating a proper word to complete the action. For each parameterized operator $M$, the probability of generating $\hat{w}$ is
\begin{equation}
    M(\hat{w}|\bm{x}, i)=\mathop{\textrm{softmax}_{\hat{w}}}(\bm{W}\bm{h}_{i})
\end{equation}
Notably, for each $M$ we train two sets of parameters for $s_{1} \rightarrow s_{2}$ and $s_{2} \rightarrow s_{1}$. For readability, we omit the direction subscripts and assure that they can be inferred from contexts; parameters of the opposite direction are denoted as $\phi_{1}'$, $\phi_{2}'$, and $\phi_{3}'$.

\subsection{Hierarchical Policy Learning}
\label{one-step}
\begin{figure}[t]
\centering
\includegraphics[width=1.0\linewidth]{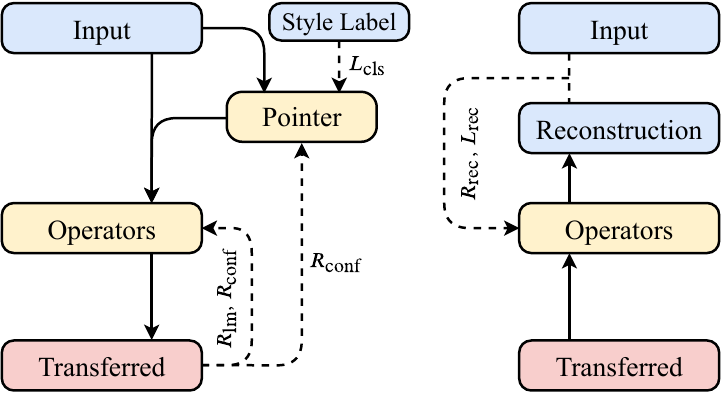}
\caption{Graphical overview for the training algorithm, which consists of a \textit{transfer} step (left) and a \textit{reconstruction} step (right). Solid lines denote forward pass; dotted lines denote rewards or losses. Blue / red items belong to the source / target styles; yellow items denotes the agents. Best viewed in color.}
\label{fig:train-overview}
\end{figure}

We introduce comprehensive training objectives to model the key aspects in text style transfer, i.e., fluency, style polarity, and content preservation. For fluency, we use an extrinsic \textit{language model reward}; for style polarity, we use an extrinsic \textit{classification confidence reward} and incorporate an auxiliary \textit{style classification} task; for content preservation, we use a self-supervised \textit{reconstruction loss} and an intrinsic \textit{reconstruction reward}. In the following parts, we only illustrate equations related to $\bm{x}_{1} \rightarrow \hat{\bm{x}}_{2}$ operations and $\hat{\bm{x}}_{2} \rightarrow \bm{x}_{1}$ reconstructions for brevity; the opposite direction can be derived by swapping $1$ and $2$. The training algorithm is presented in Algorithm~\ref{alg:one-step}. A graphical overview is shown in Figure~\ref{fig:train-overview}.
\begin{algorithm}[tb]
\small
   \caption{Point-Then-Operate Training}
   \label{alg:one-step}
\begin{algorithmic}[1]
   \STATE {\bfseries Input:} Non-aligned sets of sentences $\mathcal{X}_{1,2}$
   \STATE Initialize $\theta$, $\phi_{1,2,3}$
   \STATE Train language models $\textrm{LM}_{2}$ on $\mathcal{X}_{2}$
   \STATE Pre-train $\theta$ by optimizing $\mathcal{L}_{\textrm{cls}}^{\theta}$ \COMMENT{Eq.~\ref{cls-loss}}
   \FOR{each iteration $i=1, 2, \cdots, m$}
       \STATE Sample $\bm{x}_{1}$ from $\mathcal{X}_{1}$
       \STATE Sample $i$ from $\mu_{\theta}(i|\bm{x}_{1})$ \COMMENT{Eq.~\ref{eq:ptr}}
       \STATE Uniformly sample $M$ \COMMENT{Table~\ref{tab:operators}}
       \STATE $\hat{\bm{x}}_{2} \leftarrow \textrm{Transfer}(\bm{x}_{1}, M, i)$ \COMMENT{Table~\ref{tab:operators}}
       \STATE Compute $R_{\textrm{conf}}$ and $R_{\textrm{lm}}$ \COMMENT{Eq.~\ref{reward-lm} and \ref{reward-conf}}
       \STATE Update $\theta$ based on $\mathcal{L}_{\textrm{cls}}^{\theta}$ and $\nabla_{\theta} J(\theta)$ \COMMENT{Eq.~\ref{cls-loss} and \ref{policy-pointer}}
       \STATE Get $M'$ and $i'$ \COMMENT{Table~\ref{tab:prime}}
       \IF{$M'$ is parameterized by $\phi'$}
           \STATE $\bar{\bm{x}}_{1} \leftarrow \textrm{Reconstruct}(\hat{\bm{x}}_{2}, M', i')$ \COMMENT{Table~\ref{tab:operators}}
           \STATE Update $\phi'$ by optimizing $\mathcal{L}_{\textrm{rec}}^{\phi'}$ \COMMENT{Eq.~\ref{loss-rec}}
       \ENDIF
       \IF{$M$ is parameterized by $\phi$}
           \STATE Compute $R_{\textrm{rec}}$ if $M$ is $\textrm{Rep}_{\phi_{3}}$ \COMMENT{Eq.~\ref{reward-rec}}
           \STATE Update $\phi$ with $\nabla_{\phi} J(\phi)$  \COMMENT{Eq.~\ref{policy-operator}}
       \ENDIF
   \ENDFOR
\end{algorithmic}
\end{algorithm}

\subsubsection{Modeling Fluency}
\paragraph{Language Model Reward} To improve the fluency, we adopt a language model reward. Let $\textrm{LM}_{1}$, $\textrm{LM}_{2}$ denote the language models for $s_{1}$ and $s_{2}$, respectively. Given the generated word $\hat{w}$ in the operated sentence $\hat{\bm{x}}_{2}$, the language model reward is defined as
\begin{equation}
\label{reward-lm}
    R_{\textrm{lm}} = \lambda_{\textrm{lm}} \textrm{LM}_{2}(\hat{w}|\hat{\bm{x}}_{2})
\end{equation}
where $\textrm{LM}_{2}(\hat{w}|\hat{\bm{x}}_{2})$ denotes the probability of $\hat{w}$ given \textit{other} words in $\hat{\bm{x}}_{2}$. In our experiments, the probability is computed by averaging a forward LSTM-LM and a backward LSTM-LM.

\subsubsection{Modeling Style Polarity}
\label{auxiliary-task-style-classification}
\paragraph{Classification Confidence Reward} We observe that language models are not adequate to capture style polarity; thus, we encourage larger change in the confidence of a style classifier, by adopting a classification confidence reward, i.e.,
\begin{equation}
\label{reward-conf}
    R_{\textrm{conf}}=\lambda_{\textrm{conf}}[p(s_{2}|\hat{\bm{x}}_{2}) - p(s_{2}|\bm{x}_{1})]
\end{equation}
where we reuse the classifier defined in Eq.~\ref{cls-softmax}.
\paragraph{Auxiliary Task: Style Classification} In HRL, the high-level policy usually suffers from the high variance of gradients since the estimated gradients are dependent on the poorly trained low-level policy. To stabilize the high-level policy learning, we introduce auxiliary supervision to the pointer. Specifically, we extend the pointer to an attention-based classifier, i.e.,
\begin{equation}
\label{cls-softmax}
    p(s_{j}|\bm{x}) = \mathop{\textrm{softmax}_{j}}(\bm{W}\sum_{i=1}^{T}\mu(i|\bm{x})\bm{h}_{i})
\end{equation}
for $j = 1, 2$. Let $\theta$ denotes the parameters of the pointer. The auxiliary classification loss for $\theta$ is
\begin{equation}
\label{cls-loss}
    \mathcal{L}_{\textrm{cls}}^{\theta}=\sum_{j=1,2}\mathbb{E}_{\bm{x}_{j} \sim \mathcal{X}_{j}}[-\log p_{\theta}(s_{j}|\bm{x}_{j})]
\end{equation}
The underlying assumption is that positions with larger attention weights for classification are more likely to be critical to style transfer.

\subsubsection{Modeling Content Preservation}
\paragraph{Self-Supervised Reconstruction Loss} To improve content preservation, we propose a reconstruction loss that guides the operator modules with self-supervision.
\begin{table}[t]
\centering
\small
\begin{tabular}{@{}lcc@{}}
\toprule
$\bm{M}$ & $\bm{M'}$ & $\bm{i'}$ \\
\midrule
$\textrm{Rep}_{\phi_{3}}$ & $\textrm{Rep}_{\phi_{3}'}$ & $i$ \\
$\textrm{DC}$ & $\textrm{IF}_{\phi_{1}'}$ or $\textrm{IB}_{\phi_{2}'}$ & $i$ or $i-1$ \\
$\textrm{DF}$ & $\textrm{IF}_{\phi_{1}'}$ or $\textrm{IB}_{\phi_{2}'}$ & $i-1$ or $i-2$ \\
$\textrm{DB}$ & $\textrm{IF}_{\phi_{1}'}$ or $\textrm{IB}_{\phi_{2}'}$ & $i+1$ or $i$ \\
\bottomrule
\end{tabular}
\caption{\label{tab:prime} Construction of self-supervised data.}
\end{table}
Suppose the word $w$ at the $i^{th}$ position is deleted or replaced by operator $M$, we identify the reconstruction operator $M'$ and reconstruction position $i'$ in Table~\ref{tab:prime}. Then $M'$ is updated with MLE, by operating on position $i'$ in $\hat{\bm{x}}_{2}$ with $w$ as gold output. For those with two ($M'$, $i'$) pairs, we uniformly sample one for training. Formally, the reconstruction loss is defined as
\begin{equation}
\label{loss-rec}
    \mathcal{L}^{\phi'}_{\textrm{rec}}=-\log M'(w|\hat{\bm{x}}_{2}, i')
\end{equation}
\paragraph{Reconstruction Reward} One-to-one transfer (e.g., $\{$\textit{delicious$\leftrightarrow$bland}, \textit{caring$\leftrightarrow$unconcerned}$\}$) is usually preferable to many-to-one transfer (e.g., $\{$\textit{delicious$\rightarrow$bad}, \textit{caring$\rightarrow$bad}$\}$). Thus, we introduce a reconstruction reward for $\textrm{Rep}_{\phi_{3}}$ to encourage one-to-one transfer, i.e.,
\begin{equation}
\label{reward-rec}
    R_{\textrm{rec}} =- \lambda_{\textrm{rec}}\mathcal{L}^{\phi_{3}'}_{\textrm{rec}}
\end{equation}
where $\mathcal{L}^{\phi_{3}'}_{\textrm{rec}}$ is the reconstruction loss in Eq.~\ref{loss-rec}.

\subsubsection{Training with Single-Option Trajectory}
\label{objectives}
Instead of executing multi-option trajectories, we only allow the high-level agent to execute a single option per episode during training, and leave the multi-option scenario to the inference algorithm (\S\ref{multi-step}). We have two motivations for executing single-option trajectories: 1) executing multi-option trajectories is less tractable and stable, especially in the case of style transfer which is sensitive to nuances in the sentence; 2) self-supervised reconstruction is ambiguous in a multi-option trajectory, i.e., the gold trajectory for reconstruction is not deterministic.

\paragraph{High-Level Policy Gradients} Since the language model reward is more \textit{local} and increases the variance of estimated gradients, we only use the classification confidence reward for the high-level policy. The policy gradient is
\begin{equation}
\label{policy-pointer}
    \nabla_{\theta} J(\theta) = \mathbb{E}_{i}[R_{\textrm{conf}}\cdot\nabla_{\theta}\log \mu_{\theta}(i|\bm{x}_{1})]
\end{equation}
where gradients are detached from $R_{\textrm{conf}}$.

\paragraph{Low-Level Policy Gradients} All the extrinsic and intrinsic rewards are used for low-level policy learning. Specifically, the rewards for $\phi_{1,2,3}$ are
\begin{equation}
\begin{split}
    & R_{1,2} = R_{\textrm{lm}} + R_{\textrm{conf}} \\
    & R_{3} = R_{\textrm{lm}} + R_{\textrm{conf}} + R_{\textrm{rec}}
\end{split}
\end{equation}
For $\phi=\phi_{1}$, $\phi_{2}$, $\phi_{3}$, the policy gradient is
\begin{equation}
\label{policy-operator}
\begin{split}
    & \nabla_{\phi} J(\phi) = \mathbb{E}_{\hat{w}}[R\cdot\nabla_{\phi}\log M_{\phi}(\hat{w}|\bm{x}_{1}, i)] \\
\end{split}
\end{equation}
\paragraph{Overall Objectives} The overall objectives for $\theta$ are the classification loss in Eq.~\ref{cls-loss} and the policy gradient in Eq.~\ref{policy-pointer}. The overall objectives for $\phi_{1,2,3}$ are the reconstruction loss in Eq.~\ref{loss-rec} and the policy gradients in Eq.~\ref{policy-operator}.

\subsection{Inference}
\label{multi-step}
The main problems in applying single-step trained modules to the multi-step scenario are 1) previous steps of operations may influence later steps, and 2) we need to dynamically decide when the trajectory should terminate. We leverage a mask mechanism to address these problems. The basic idea is that given an input sentence, the high-level agent iteratively proposes operation positions for the low-level agent to operate around. In each iteration, the high-level agent sees the whole sentence but with some options (i.e., positions) masked in its policy. The trajectory termination condition is modeled by an additional pre-trained classifier. The algorithm for style transfer from $s_{1}$ to $s_{2}$ is detailed in Algorithm \ref{alg:multi-step-inference}. 
\begin{algorithm}[tb]
\small
   \caption{Point-Then-Operate Inference}
   \label{alg:multi-step-inference}
\begin{algorithmic}[1]
   \STATE {\bfseries Input:} Input sentence $\bm{x}_{1}$, additional classifier $p_{\textrm{add}}$
   \STATE Initialize $\hat{\bm{x}}_{2} \leftarrow \bm{x}_{1}$, $\hat{\bm{x}}_{2}^{m} \leftarrow \bm{x}_{1}$, $j \leftarrow 1$
   \WHILE{$p_{\textrm{add}}(s_{1}|\hat{\bm{x}}_{2}^{m})>p_{\textrm{stop}}$ and $j \leq j_{\max}$}
       \STATE Mask the options in $\mu_{\theta}(i|\hat{\bm{x}}_{2})$ \COMMENT{\S\ref{masked-options}}
       \STATE Select $i$ that maximizes the masked $\mu_{\theta}(i|\hat{\bm{x}}_{2})$ 
       \STATE Select the \textit{best} $M$ from Table~\ref{tab:operators} \COMMENT{\S\ref{choice-of-operator-modules}}
       \STATE Update $\hat{\bm{x}}_{2} \leftarrow \textrm{Transfer}(\hat{\bm{x}}_{2}, M, i)$ \COMMENT{\S\ref{operators}}
       \STATE Update $\hat{\bm{x}}_{2}^{m}$ \COMMENT{\S\ref{masked-transferred}}
       \STATE $j \leftarrow j+1$
   \ENDWHILE
\STATE The output is $\bm{x}_{1 \rightarrow 2} \leftarrow \hat{\bm{x}}_{2}$
\RETURN $\bm{x}_{1 \rightarrow 2}$
\end{algorithmic}
\end{algorithm}
 
\subsubsection{Masked Options}
\label{masked-options}
To tackle the first problem, we mask the options (i.e., positions) in the high-level policy which appear in the contexts in which any words are inserted, replaced, or skipped (but not for deleted words). Note that we only mask the options in the policy but do \textit{not} mask the words in the sentence (i.e., both agents still receive the complete sentence), since we cannot bias the state representations (\S\ref{pointer} and \S\ref{operators}) with masked tokens. We set the window size as 1 (i.e., three words are masked in each step). We find the use of window size necessary, since in many cases, e.g., negation and emphasis, the window size of 1 is capable of covering a complete semantic unit.

\subsubsection{Termination Condition}
\label{masked-transferred}
A simple solution to the second problem is to terminate the trajectory if the operated sentence is confidently classified as the target style. The problem with this simple solution is that the highly stylized part may result in too early termination. For example, \textit{Otherwise a terrible experience and we \textbf{will} go again} may be classified as negative with high confidence. Thus, we propose to mask words in the operated sentence for the termination condition. The masking strategy is the same as \S\ref{masked-options} and masked words are replaced by $\langle\textrm{unk}\rangle$. To tackle the excessive number of $\langle\textrm{unk}\rangle$, we train an additional classifier as defined in \S\ref{auxiliary-task-style-classification}, but trained on sentences with words randomly replaced as $\langle\textrm{unk}\rangle$.

\subsubsection{Inference Policy for Operator Selection}
\label{choice-of-operator-modules}
As discussed in \S\ref{operators}, we adopt a heuristic inference policy for operator selection. Specifically, we enumerate each operator and select the operated sentence $\hat{\bm{x}}_{2}$ which maximizes the criterion:
\begin{equation}
\label{criterion}
    c(\hat{\bm{x}}_{2})=\textrm{LM}_{2}(\hat{\bm{x}}_{2})\cdot p(s_{2}|\hat{\bm{x}}_{2})^{\eta}
\end{equation}
where $\textrm{LM}_{2}(\hat{\bm{x}}_{2})$ denotes the probability of $\hat{\bm{x}}_{2}$ computed by the language model $\textrm{LM}_{2}$, $p(s_{j}|\cdot)$ is the classifier defined in \S\ref{auxiliary-task-style-classification}, and $\eta$ is a balancing hyper-parameter.

\section{Experiments}
\subsection{Datasets}
\label{datasets}
We conduct experiments on two commonly used datasets for unsupervised text style transfer, i.e., Yelp and Amazon, following the split of datasets in \citet{li2018delete}. Dataset statistics are shown in Table~\ref{tab:datasets-stats}. For each dataset, \citet{li2018delete} provided a gold output for each entry in the test set written by crowd-workers on Amazon Mechanical Turk. Since gold outputs are \textit{not} written for development sets, we tune the hyper-parameters on the development sets based on our intuition of English.  
\paragraph{Yelp} The Yelp dataset consists of business reviews and their labeled sentiments (from 1 to 5) from Yelp. Those labeled greater than $3$ are considered as positive samples and those labeled smaller than $3$ are negative samples.
\paragraph{Amazon} The Amazon dataset consists of product reviews and labeled sentiments from Amazon \cite{he2016ups}. Positive and negative samples are defined in the same way as Yelp.

We observe that the Amazon dataset contains many neutral or wrongly labeled sentences, which greatly harms our HRL-based sequence operation method. Thus, on the Amazon dataset, we adopt a \textit{cross-domain} setting, i.e., we train the modules on the Yelp training set using the Amazon vocabulary and test the method on Amazon test set. Experimental results show the effectiveness of our method under this cross-domain setting.

\begin{table}[t]
\small
\centering
\begin{tabular}{@{}lcrrr@{}}
\toprule
\bf Dataset & \bf Attributes & \bf Train & \bf Dev & \bf Test \\\midrule
 \multirow{2}*{Yelp} & Positive & 270K & 2000 & 500 \\
 ~ & Negative & 180K & 2000 & 500 \\ \midrule
 \multirow{2}*{Amazon} & Positive & 277K & 985 & 500 \\
 ~ & Negative & 278K & 1015 & 500 \\
 \bottomrule
\end{tabular}
\caption{\label{tab:datasets-stats} Dataset statistics.}
\end{table}

\subsection{Evaluation Metrics}
\label{evaluation-metrics}
\paragraph{Automatic Evaluation} 
Following previous work \cite{shen2017style, xu2018unpaired}, we pre-train a style classifier TextCNN \cite{kim2014convolutional} on each dataset and measure the style polarity of system outputs based on the classification accuracy. Also, based on the human references provided by \citet{li2018delete}, we adopt a case-insensitive BLEU metric, which is computed using the Moses \texttt{multi-bleu.perl} script.
\paragraph{Human Evaluation}
Following previous work \cite{shen2017style, xu2018unpaired}, we also conduct human evaluations. For each input sentence and corresponding output, each participant is asked to score from 1 to 5 for fluency, content preservation, and style polarity. If a transfer gets scores of 4 or 5 on all three aspects, it is considered as a successful transfer. We count the success rate over the test set for each system, which is denoted as \textit{Suc} in Table~\ref{tab:human-results}.

\subsection{Baselines}
We make a comprehensive comparison with state-of-the-art style transfer methods. \textbf{CrossAligned} \cite{shen2017style} aligns decoder hidden states adversarially. \textbf{MultiDecoder} \cite{fu2018style} adopts multiple decoders for different styles. \textbf{StyleEmbedding} \cite{fu2018style} adopts a single decoder conditioned on learned style embeddings. \textbf{TemplateBased} \cite{li2018delete} retrieves and replaces stylized words. \textbf{DeleteOnly} \cite{li2018delete} only deletes the stylized words in the input sentence. \textbf{Del-Ret-Gen} \cite{li2018delete} is the same as TemplateBased except that an RNN is adopted to generate the output. \textbf{BackTranslate} \cite{prabhumoye2018style} stylizes the back-translated input. \textbf{UnpairedRL} \cite{xu2018unpaired} deletes stylized words and generates with a denoising AE. \textbf{UnsuperMT} \cite{zhang2018style} produces pseudo-aligned data and iteratively learns two NMT models. 

The outputs of the first six baselines are made public by \citet{li2018delete}. The outputs of BackTranslate and UnpairedRL are obtained by running the publicly available codes. We get the outputs of UnsuperMT from the authors of \citet{zhang2018style}.

\subsection{Evaluation Results}
\label{evaluation-results}
Table~\ref{tab:automatic-results} shows the results of automatic evaluation. It should be noted that the classification accuracy for human reference is relatively low (74.7\% on Yelp and 43.2\% on Amazon); thus, we do \textit{not} consider it as a valid metric for comparison. For BLEU score, our method outperforms recent systems by a large margin, which shows that our outputs have higher overlap with reference sentences provided by humans. 
\begin{table}[t]
\centering
\small
\begin{tabular}{@{}lrrrrrr@{}}
\toprule
\multirow{2}*{} & \multicolumn{2}{c}{\bf Yelp} & \multicolumn{2}{c}{\bf Amazon} \\
\cmidrule(lr){2-3} \cmidrule(l){4-5}
~                    & Acc     & BLEU       & Acc     & BLEU      \\
\midrule
CrossAligned         &74.7     &9.06        &75.1     &1.90       \\
MultiDecoder         &50.6     &14.54       &69.9     &9.07       \\
StyleEmbedding       &8.4      &21.06       &38.2     &15.07      \\
TemplateBased        &81.2     &22.57       &64.3     &34.79      \\
DeleteOnly           &86.0     &14.64       &47.0     &33.00      \\
Del-Ret-Gen          &88.6     &15.96       &51.0     &30.09      \\
BackTranslate      &94.6     &2.46        &\bf 76.7 &1.04       \\
UnpairedRL           &57.5     &18.81       &56.3     &15.93      \\
UnsuperMT            &\bf 97.8 &22.75       &72.4     &33.95      \\
\midrule
Human       &74.7     &-           &43.2     &-          \\
\midrule
Point-Then-Operate   &91.5     &\bf 29.86   &40.2     &\bf 41.86  \\
\bottomrule
\end{tabular}
\caption{\label{tab:automatic-results} Automatic evaluation results for classification accuracy and BLEU with human reference. \textit{Human} denotes human references. Note that Acc for human references are relatively low; thus, we do \textit{not} consider it as a valid metric for comparison.}
\end{table}
\begin{table*}[t]
\small
\centering
\begin{tabular}{@{}lcccrcccr@{}}
\toprule
\multirow{2}*{} & \multicolumn{4}{c}{\bf Yelp} & \multicolumn{4}{c}{\bf Amazon} \\
\cmidrule(lr){2-5} \cmidrule(l){6-9}
~                 & Fluency & Content  & Style & \multicolumn{1}{c}{Suc} & Fluency    & Content  & Style  & \multicolumn{1}{c}{Suc} \\
\midrule
TemplateBased      & 3.47      & 3.76         & 3.25       & 68.0 \% & 3.46   & 4.08    & 2.15      & 9.0 \%  \\
Del-Ret-Gen        & 3.82      & 3.73         & 3.52       & 70.3 \% & 4.02   & 4.31    & 2.69       & 21.0 \% \\
UnpairedRL         & 3.54      &  3.59        & 2.90       & 53.8  \% & 2.58   &  2.55   & 2.44      & 4.5 \%   \\
UnsuperMT          & 4.26      &  4.24        & \bf 4.03       & \bf 82.5 \% & 4.24   & 4.13    & 3.05 & 35.5 \% \\
\midrule
Point-Then-Operate &\bf 4.39      & \bf 4.56         &  3.78      & 81.5 \% &\bf 4.28    & \bf 4.47  & \bf 3.31 & \bf 47.0 \% \\
\bottomrule
\end{tabular}
\caption{\label{tab:human-results} Human evaluation results. Methods are selected based on automatic evaluation. \textit{Style}: style polarity; \textit{Content}: content preservation; \textit{Fluency}: fluency; \textit{Suc}: the proportion of successful transfer (refer to \S\ref{evaluation-metrics})}
\end{table*}

To lighten the burden on human participants, we compare our proposed method to only four of the previous methods, selected based on their performance in automatic evaluation. Given the observation discussed in \S\ref{datasets}, we remove the wrongly labeled test samples for human evaluation. Table~\ref{tab:human-results} shows the results of human evaluation. Our proposed method achieves the highest fluency and content preservation on Yelp and performs the best on all human evaluation metrics on Amazon. 

\subsection{Controllable Trade-Off}
\begin{figure}[t]
\centering
\subfigure[Yelp]{
\includegraphics[width=0.46\linewidth]{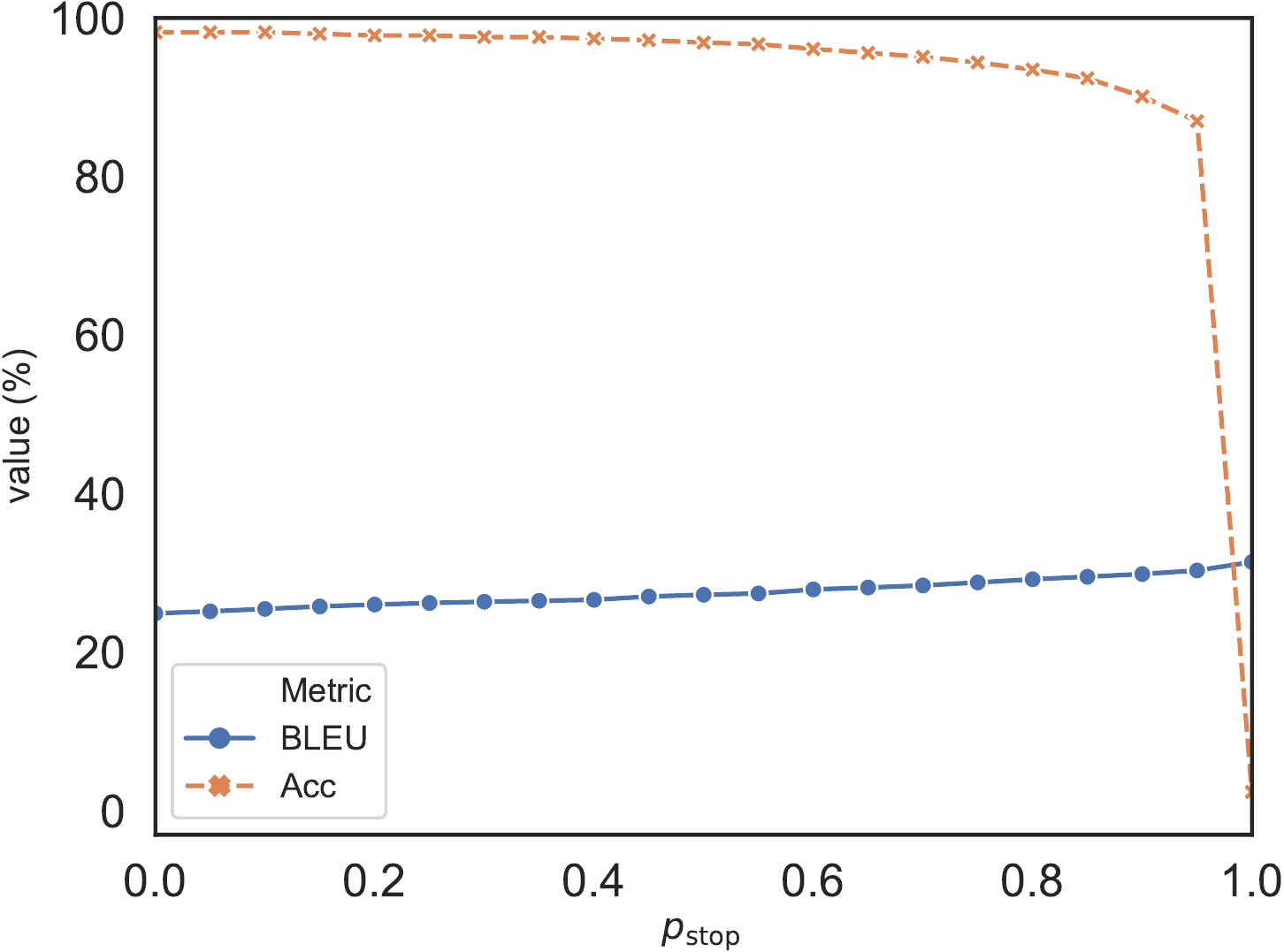}
}
\hspace{0cm}
\subfigure[Amazon]{
\includegraphics[width=0.46\linewidth]{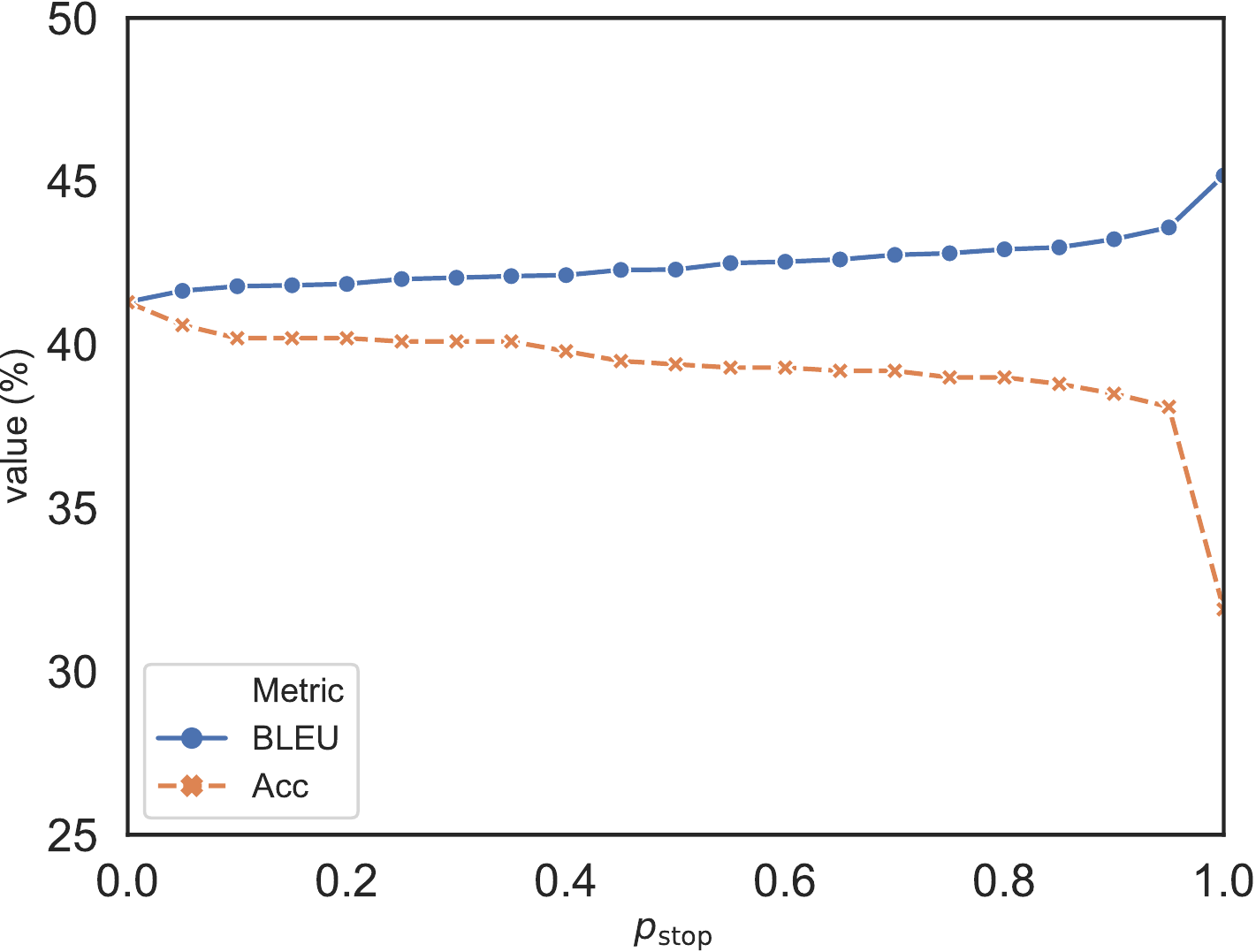}
}
\caption{The controllable trade-off between content preservation and style polarity. The $x$-axis is $p_{\textrm{stop}}$ (defined in Algorithm~\ref{alg:multi-step-inference}). The $y$-axis is the value of different automatic metrics, i.e., BLEU (the blue lines) and classification accuracy (the orange lines).}
\label{fig:p-stop}
\end{figure}

Figure~\ref{fig:p-stop} shows how classification accuracy and BLEU change when we manually set $p_{\textrm{stop}}$. When $p_{\textrm{stop}}$ is larger, classification accuracy drops and BLEU increases. Based on our observation of human references, we find that humans usually make minimal changes to the input sentence; thus, BLEU computed with human references can be viewed as an indicator of content preservation. From this perspective, Figure~\ref{fig:p-stop} shows that if we stop earlier, i.e., when the current style is closer to the source style, more content will be preserved and more weakly stylized words may be kept. Thus, controllable trade-off is achieved by manually setting $p_{\textrm{stop}}$.

\subsection{Ablation Studies}
We conduct several ablation studies to show the effect of different components in our method:
\paragraph{Ablations of Operators} To show that incorporating various operators is essential, we evaluate the performance of the following ablations: InsertOnly, ReplaceOnly, and DeleteOnly, in which operator choices are restricted to subsets of Table~\ref{tab:operators}.
\paragraph{Ablation of Reconstruction Reward and Reconstruction Loss} To show the effectiveness of our reconstruction-based objectives, we remove the reconstruction reward and the reconstruction loss as an ablation.
\begin{table}[t]
\small
\centering
\begin{tabular}{@{}lrrrr@{}}
\toprule
\multirow{2}*{} & \multicolumn{2}{c}{\bf Yelp} & \multicolumn{2}{c}{\bf Amazon} \\
\cmidrule(lr){2-3} \cmidrule(l){4-5}
~                                            & Acc     & BLEU    & Acc     & BLEU    \\
\midrule
InsertOnly                                   & 68.6    & 23.93   & 48.2    & 36.77   \\
ReplaceOnly                                  & \bf 93.8    & 26.41   & \bf 47.8    & 37.39   \\
DeleteOnly                                   & 37.6    & 25.70   & 25.0    & 41.68   \\
\midrule
w/o $R_{\textrm{rec}}$ and $L_{\textrm{rec}}$& 39.1    & 27.80   & 46.3    & 40.52   \\
\midrule
Human       &74.7     &-           &43.2     &-          \\
\midrule
Full                                         &91.5     & \bf 29.86 & 40.2    & \bf 41.86   \\
\bottomrule
\end{tabular}
\caption{\label{tab:ablation} Ablation Studies.}
\end{table}

Table~\ref{tab:ablation} shows the ablation results. It shows that BLEU drops if operators are restricted to a fixed set, showing the necessity of cooperating operator modules. It also shows that BLEU drops if we remove the reconstruction loss and the reconstruction reward, indicating the generated words overlap less with human references in this ablation case. As discussed in \S\ref{evaluation-results}, we ignore Acc since it is low on human references.

\subsection{Qualitative Study}
Figure~\ref{fig:example} is an example of our method applied to a test sample. The transfer starts from more stylized parts and ends at less stylized parts, while keeping neutral parts intact. It also shows that our method learns \textit{lexical substitution} and \textit{negation} in an unsupervised way. Table~\ref{tab:sampled-outputs} displays some comparisons of different systems. It shows that our proposed method is better at performing local changes to reverse the style of the input sentence while preserving most style-independent parts. 
\begin{table*}[t]
\centering
\small
\begin{tabular}{@{}ll@{}}

\toprule
 Original (Yelp, negative)   & staffed primarily by teenagers that do n't understand customer service . \\ 
 \midrule
 TemplateBased         & staffed primarily by teenagers that \blue{huge portions and} customer service \red{are pretty good} .\\ 
 Del-Ret-Gen           & staffed \blue{, the} \red{best} \blue{and sterile} by \blue{flies} , how \red{fantastic} customer service .\\ 
 UnpairedRL            & staffed \blue{established each tech feel when} \red{great} customer service \red{professional} .\\ 
 UnsuperMT             & staffed \blue{distance} that \red{love} customer service .\\ 
 \midrule
 Point-Then-Operate    & staffed by \red{great} teenagers that do \red{delightfully} understand customer service .\\ 
 \bottomrule
\toprule
Original (Yelp, positive) & i will be going back and enjoying this great place !\\ 
\midrule
 TemplateBased         & \blue{i will be going back} and \blue{enjoying} this \blue{i did not @unk}\\ 
 Del-Ret-Gen           & \blue{i will be going back} and will \red{not be returning} into this \\ 
 UnpairedRL            & \blue{i will be going back} and \blue{enjoying} this \blue{great} place .\\ 
 UnsuperMT             & i \red{wo n't} be going back and \blue{sitting this @num} .\\ 
 \midrule
 Point-Then-Operate    & i will \red{not} be going back and \red{avoid} this \red{horrible} place !\\ 
 \bottomrule
 \toprule
 Original (Amazon, negative) & i could barely get through it they taste so nasty .\\ 
 \midrule
 TemplateBased         & \blue{beautifully} through it they taste so \blue{nasty} .\\ 
 Del-Ret-Gen           & i \blue{have used} it through and it is very \blue{sharp} and it was very \blue{nasty} .\\ 
 UnpairedRL            & i could \blue{barely} get through it they taste so \blue{nasty} .\\ 
 UnsuperMT             & i can \blue{perfect} get through it they taste so \red{delicious} .\\ 
 \midrule
 Point-Then-Operate    & i could get through it they taste so \red{good} .\\ 
 \bottomrule
 \toprule
 Original (Amazon, positive) & i also prefered the blade weight and thickness of the wustof .\\
 \midrule
 TemplateBased         & i \blue{also} prefered the blade weight and thickness of the wustof \blue{toe} .\\ 
 Del-Ret-Gen           & i \blue{also} prefered the blade and was very \red{disappointed} in the weight and thickness of the wustof .\\ 
 UnpairedRL            & i \blue{also} \blue{sampled} the \blue{comfortable base} and \blue{follow} of the \blue{uk} .\\ 
 UnsuperMT             & i \blue{also} \blue{encounter} the blade weight and \blue{width} of the \blue{guitar} .\\ 
 \midrule
 Point-Then-Operate    & i \red{only} prefered the weight and thickness of the wustof .\\ 
 \bottomrule
 
\end{tabular}
\caption{\label{tab:sampled-outputs} Sampled system outputs. The dataset and the original style for each input sentence are parenthesized. We mark improperly generated or preserved words in \blue{blue}, and mark words that show target style and are grammatical in the context in \red{red}. Best viewed in color.}
\end{table*}

\section{Discussions}
We study the system outputs and observe two cases that our method cannot properly handle:
\paragraph{Neutral Input} The \textit{reconstruction} nature of our method prefers stylized input to neutral input. We observe that it fails to convert some neutral inputs, e.g., \textit{I bought this toy for my daughter about @num months ago.}, which shows that the high-level policy is not well learned for some neutral sentences.
\paragraph{Adjacent Stylized Words} We introduce a window size of 1 in \S\ref{masked-options} to deal with most semantic units. However, we observe in some cases two adjacent stylized words occur, e.g., \textit{poor watery food}. If the first step is to replace one of them, then the other will be masked in later iterations, leading to incomplete transfer; if the first step is deletion, our method performs well, since we do not mask the context of deletion as stated in \S\ref{masked-options}. Notably, phrases like \textit{completely horrible} is not one of these cases, since \textit{completely} itself is not stylized.

Experiments in this work show the effectiveness of our proposed method for positive-negative text style transfer. Given its sequence operation nature, we see potentials of the method for other types of transfers that require \textit{local} changes, e.g., polite-impolite and written-spoken, while further empirical verification is needed.

\section{Conclusions}
We identify three challenges of existing seq2seq methods for unsupervised text style transfer and propose Point-Then-Operate (PTO), a sequence operation-based method within the hierarchical reinforcement learning (HRL) framework consisting of a hierarchy of agents for pointing and operating respectively. We show that the key aspects of text style transfer, i.e., fluency, style polarity, and content preservation, can be modeled by comprehensive training objectives. To make the HRL training more stable, we provide an efficient mask-based inference algorithm that allows for single-option trajectory during training. Experimental results show the effectiveness of our method to address the challenges of existing methods.

\section*{Acknowledgments}
We would like to thank the anonymous reviewers for their thorough and helpful comments. We are grateful to the authors of \citet{zhang2018style} for providing the UnsuperMT results. Xu Sun is the corresponding author of this paper.
\clearpage

\bibliography{acl2019}
\bibliographystyle{acl_natbib}

\appendix



\end{document}